\documentclass[10pt,twocolumn,letterpaper]{article}
\pdfoutput=1
\usepackage{cvpr}
\usepackage{times}
\usepackage{epsfig}
\usepackage{graphicx}
\usepackage{amsmath}
\usepackage{amssymb}
\usepackage[nocompress]{cite}

\usepackage[breaklinks=true,bookmarks=false]{hyperref}
\cvprfinalcopy 

\def\cvprPaperID{****} 

\begin{document}

\title{FusiformNet: Extracting Discriminative Facial Features on Different Levels}

\author{Kyo Takano\\
Independent Researcher\\
United States \& Japan \\
{\tt\small kyo.psych@gmail.com}\\
}

\maketitle

\begin{abstract}
Over the last several years, research on facial recognition based on Deep Neural Network has evolved with approaches like task-specific loss functions, image normalization and augmentation, network architectures, etc. However, there have been few approaches with attention to how human faces differ from person to person. Premising that inter-personal differences are found both generally and locally on the human face, I propose FusiformNet, a novel framework for feature extraction that leverages the nature of discriminative facial features. Tested on Image-Unrestricted setting of Labeled Faces in the Wild benchmark, this method achieved a state-of-the-art accuracy of 96.67\% without labeled outside data, image augmentation, normalization, or special loss functions. Likewise, the method also performed on a par with previous state-of-the-arts when pre-trained on CASIA-WebFace dataset. Considering its ability to extract both general and local facial features, the utility of FusiformNet may not be limited to facial recognition but also extend to other DNN-based tasks.

\end{abstract}

\section{Introduction}
Facial recognition is an automated recognition of individuals’ facial images, and it involves DNN-based extraction and comparison of discriminative facial features that are independent of extraneous factors such as pose, expression, makeup, lighting, image quality, etc. As Wang \& Deng (2020) \cite{wang2020deep} summarized in their comprehensive survey, today’s facial recognition technology has been advanced to a super-human performance through approaches like the following: loss functions, DNN-based normalization \& augmentation, and network architectures, etc. 

Paying specific attention to the architecture-based approach, the mainstream deep facial recognition systems rely on a simple DNN that takes only a single image to extract a vector representation of a face \cite{wang2020deep}. On the other hand, researchers have also explored different architectures for facial recognition that effectively make the most of the characteristics of the data, or the human face. One remarkable approach is to feed multiple images of a face from different angles; by doing so, a DNN can implicitly grasp the 3D structure of a face when extracting pose-invariant features. Masi \etal (2016) \cite{masi2016pose} proposed Pose-Aware Models, in which latent vectors are extracted from different DNNs corresponding to specific angles, instead of feeding an image with large pose variability into a single DNN. Similarly, MvDN proposed by Kan \etal \cite{kan2016multi} forced its pose-specific sub-networks to extract only pose-invariant facial features. This approach is conceptually similar to image augmentation with GAN-based face rotation (CAPG-GAN\cite{hu2018pose}, DA-GAN \cite{zhao2017dual}, DR-GAN \cite{tran2017disentangled}), in that 3D physical structures are implicitly inferred when rotating/frontalizing a face. However, although feature extractors used in generating frontal/rotated faces can extract pose-invariant features, such information is lost once the image is processed before being fed into a subsequent DNN as a training sample.

Although the pose-invariant feature extraction with multi-view inputs is an effective approach in that it leverages the three-dimensional nature of the human face, researchers have not still paid sufficient attention to detailed characteristics of discriminative facial features. To address this point, I premise that the human face varies roughly on two different levels: general composition and local and subtle features. While human faces share the basic components---skin, eyes, nose, mouth, etc.---with a similar composition, there are variations in their sizes, shapes, coordinates, colors, etc; besides such general variances, inter-personal variances can also be found locally on more specific details, such as eye-angles, skin quality, etc. This distinction is analogous with when you find your friend from a distance and when you confirm it up close.

Based on the premise, I propose FusiformNet, a simple and robust framework that effectively extracts discriminative features from a single input image without training on labeled data. FusiformNet outperformed the current state-of-the-art result of \textit{Labeled Faces in the Wild} (\textit{LFW}) benchmark on \textit{Label-Free Outside Data} (\textit{Label-Free}) setting without image normalization, augmentation, or special loss functions. Also, when pre-trained on \textit{CASIA-WebFace} dataset, the method performed almost on a par with state-of-the-art results on \textit{Labeled Outside Data} (\textit{Labeled}) setting of LFW. Let alone its high performance on the facial recognition task, the scheme of FusiformNet is very intuitive and easy to implement.

\section{Proposed Method}
FusiformNet is a framework of facial feature extraction utilizing a bottleneck autoencoder and a feature extractor. In this section, first, I briefly introduce the concepts of the bottleneck autoencoder and the feature extractor. Then, the concept of FusiformNet based on these two DNNs is explained.

\subsection{Bottleneck Autoencoder}
A bottleneck autoencoder comprises an encoder and a decoder. Encoder extracts essential data through convolution and compresses it into a vector representation at the bottleneck. Subsequently, decoder reconstructs the image from the vector representation through de-convolution layers. Theoretically, image autoencoding functions as dimension reduction when the vector representation is retrieved from the bottleneck. Given the utility of dimension reduction, autoencoder had been used in initializing weights \cite{erhan2010does} until researchers had later found random weight initialization is good enough to train DNNs on labeled data \cite{chollet2016building}. This rejection of autoencoder despite its compression function can be explained by huge perceptual loss attributed to pixel-wise image reconstruction; pixel-wise loss does not optimize for visual features that are regionally small but perceptually significant, which several studies have stated in the past \cite{larsen2016autoencoding, snell2017learning}.

\begin{figure}[t]
\begin{center}
\includegraphics[width=1.0\linewidth]{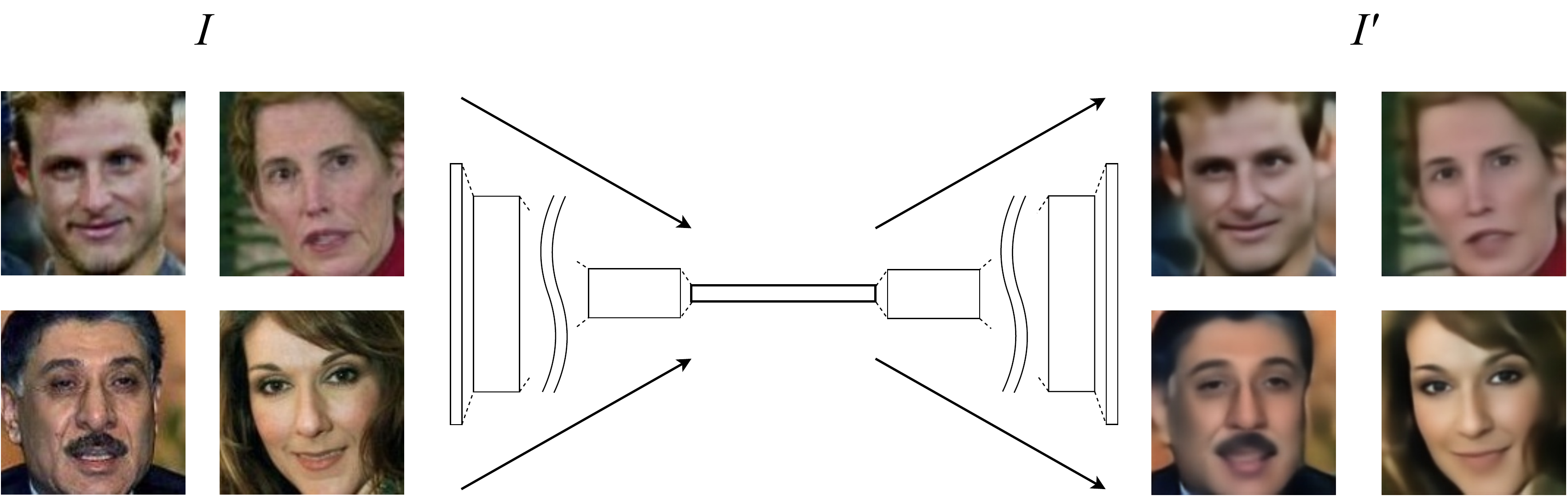}
\end{center}
   \caption{Illustrated bottleneck autoencoder. By going through a narrow bottleneck, the input image is blurred and de-identified. The sample image is from LFW dataset.}
\label{fig:autoencoder}
\end{figure}

In the present paper, I propose to paradoxically make use of the perceptual loss caused by pixel-wise reconstruction. With a very narrow bottleneck compared to amounts of information in an image, only a limited amount of information describing the face thumbnail can go through the bottleneck. As a result, input face images are de-identified through the process of reconstruction as they get \textit{closer} to more average faces while preserving its general composition. Subtle visual features like eye angles and skin qualities are blurred and normalized to look more like those of an average face because such subtle features can be considered as a sort of noise deviating from the expected distributions (see figure~\ref{fig:autoencoder} for an example). While preserving general structural information such as approximate facial angles, landmarks, and colors, images are reconstructed with somewhat normalized facial features.

The bottleneck autoencoder for the present study is trained in an unsupervised manner on cropped VGGFace2 \cite{cao2018vggface2} face thumbnails, employing Xception's base architecture \cite{chollet2017xception} for the encoder part and piling up deconvolution layers as the decoder. With a $2048$-$dimension$ bottleneck, the autoencoder’s pixel-wise Mean Square Error defined as equation~\ref{L_pixel} converged to $0.0023$.

\begin{equation} \label{L_pixel}
 L_{pixel} = \frac{1}{h \cdot w} 
\sum_{}^{h} \sum_{}^{w} \left\|I_{x, y}-R(L)_{x, y}\right\|^{2} 
\end{equation}

\subsection{Perceptual DNN}
DNNs pre-trained on the ImageNet classification task \cite{russakovsky2015imagenet} are well known for their perceptual ability on a variety of visual tasks. Following the idea of perceptual loss proposed by Johnson \etal \cite{johnson2016perceptual}, in the present study, I employed Xception pre-trained on ImageNet as a feature extractor. With all the weights freezed, the average-pooled output of the last (deepest) Xception block is obtained as a feature vector of 2048 dimensions representing the general features of an input image.

\begin{figure}[t]
\begin{center}
\includegraphics[width=1.0\linewidth]{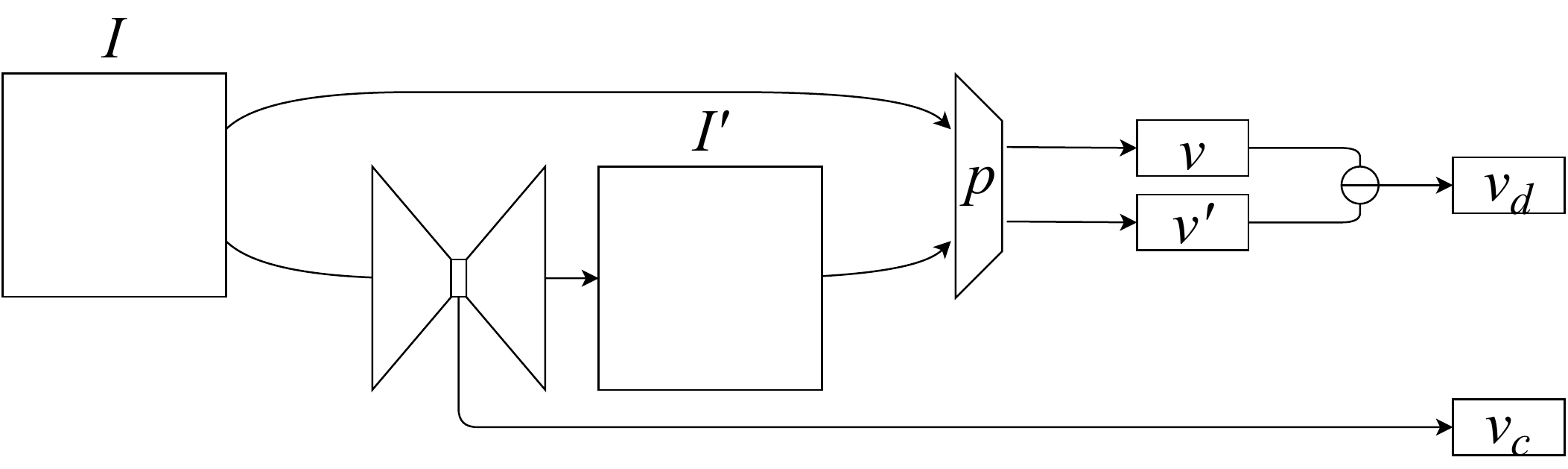}
\end{center}
   \caption{Illustrated framework of FusiformNet. Given an image $I$, FusiformNet returns $v_c$ and $v_d$ as general and local facial features, respectively. $p$ denotes the perceptual DNN and $-$ is an operator for element-wise subtraction.}
\label{fig:fusiformnet}
\end{figure}

\subsection{FusiformNet}
The idea of FusiformNet is straightforward and intuitive. As Figure~\ref{fig:fusiformnet} depicts, given a facial image, FusiformNet extracts a latent vector describing the general facial structures and another representing the perceptual difference between original and de-identified images. The network first passes an input image $I$ through the bottleneck autoencoder to get an image $I’$ that entails less person-identifying features. In this process, $v_c$, a compressed vector representation of the general structure of the input image is extracted from the bottleneck. Subsequently, by feeding each of the pair into the perceptual DNN and subtracting the two output vectors element-wise, the perceptual difference between the original and reconstructed images is distilled as $v_d$. Here, because the reconstructed image is de-identified due to the effect of pixel-wise autoencoding, the perceptual difference $v_d$ can be differentially considered as person-identifying features that make up one’s identity relative to the compressed vector $v_c$. Although those vectors are not purely tuned for facial recognition inferences, this explicit distinction between general and local levels enables comparison on each level and subsequent integration. The method is named after \textit{Fusiform Face Area}, a brain region in charge of facial recognition and the fusiform shape of the network flow caused by the autoencoder.

\section{Experiment}
The proposed FusiformNet is evaluated on the Image-Unrestricted setting of LFW verification tasks. 

\subsection{Experimental settings}
As preprocessing, all training data are cropped out of original images like MTCNN face detector does, preserving their original aspect ratios. The face crops are preprocessed to size 224 x 224 pixels in RGB and each pixel values are standardized to vary in the range [0, 1], and no data was augmented in the experiment for all conditions. 

Adam \cite{kingma2017adam} with the initial learning rate of 0.0001 was used as the weight optimizer and I set the batch size to 600. Each training session continued for 80K steps on a Nvidia Tesla P100 with Keras. While FusiformNet’s weights are freezed for all sessions, the facial verification module described below was trained.

\subsection{Facial verification module}

\begin{figure}[t]
\begin{center}
\includegraphics[width=1.0\linewidth]{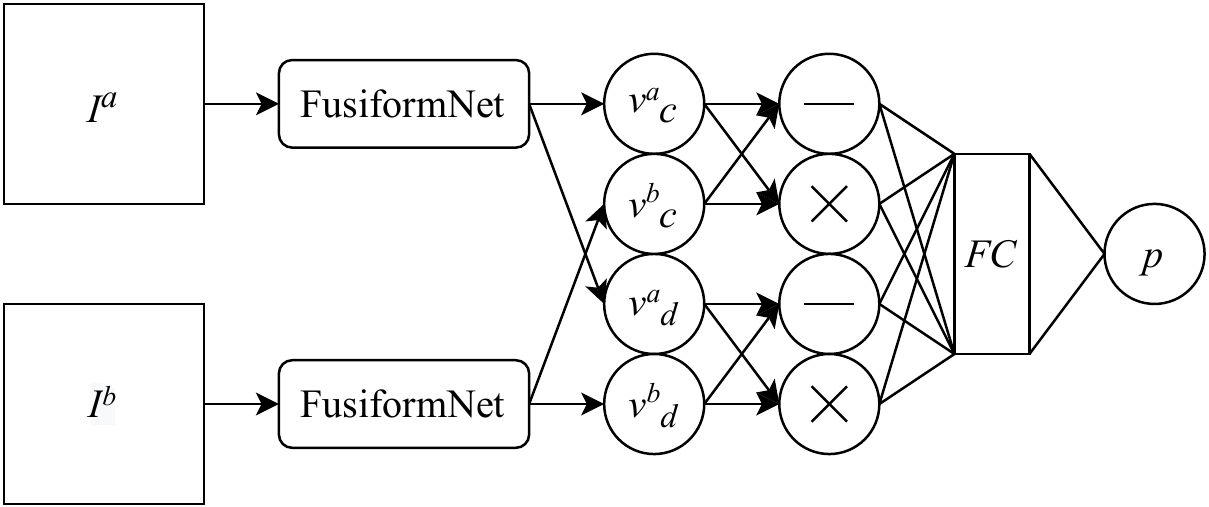}
\end{center}
   \caption{Illustrated network of the facial verification module based on FusiformNet. Images $I^a$ and $I^b$ are either of a same person or two different identities. Operators $-$ and $\times$ represents element-wise subtraction and multiplication, respectively.}
\label{fig:fr_module}
\end{figure}

In the verification task, a pair of images, either matching or unmatching, is fed into the algorithm as shown in Figure~\ref{fig:fr_module}. For both pairs of $v_c$ and $v_d$, element-wise difference and product are computed. Concatenating them and feeding to the subsequent Fully-Connected layer ($1024$-$D$) and then to the output later with the sigmoid activation, the facial verification module predicts face similarity in the range [0, 1].

\subsection{Evaluation}
FusiformNet is evaluated in both \textit{Label-Free} and \textit{Labeled} settings separately. In the Label-Free setting, no labeled external data, no data labels describing input data is allowed when pre-training the model, while the Labeled setting allows the use of labeled identity-related labels outside LFW dataset \cite{huang2014labeled}. For both settings, to repeat, the bottleneck autoencoder inside FusiformNet is trained on VGGFace2 dataset \cite{cao2018vggface2} in an unsupervised manner. For the Labeled setting, on the other hand, CASIA-WebFace dataset was used to pre-train the facial verification module that processes outputs from FusiformNet to compare two images. Following the protocol of LFW, I employed 10-fold cross-validation in evaluating different methods. I also explored performances of the pre-trained Xception and FusiformNet components, namely $v_c$ and $v_d$; when testing the performance of $v_c$ and $v_d$ respectively, the other is disabled on the facial verification module for each pair of images.

\subsection{Results}
The evaluation results are summarized in Table~\ref{table:results_0}. FusiformNet reached state-of-the-art accuracy of $96.45\%$ on Label-Free setting and a high performance on Labeled setting that is on a par with previous state-of-the-art results. While FusiformNet outperformed ImageNet pre-trained Xception in both settings, the results show that its components $v_c$ and $v_d$ alone do not necessarily perform better than Xception and previous state-of-the-art results.

\begin{table}
\begin{center}
\begin{tabular}{l|l|l}
\hline
Method & Label-Free & Labeled \\
\hline\hline
MDML-DCPs \cite{Ding_2016} & $95.58 \pm 0.34$ & \\
Frankenstein \cite{hu2017frankenstein} & \textbf{95.77 $\pm$ 0.38} & \\
DeepFace \cite{taigman2014deepface} & & $97.35 \pm 0.25$ \\
FaceNet \cite{schroff2015facenet} & & $99.63 \pm 0.10$ \\
ArcFace \cite{deng2019arcface} & & \textbf{99.83} \\
\hline\hline
\textbf{FusiformNet} & \textbf{96.45 $\pm$ 0.79} & \textbf{99.30 $\pm$ 0.15} \\
\quad $v_c$ (compression) & $94.25 \pm 0.61$ & $98.23 \pm 0.39$ \\
\quad $v_d$ ($\Delta$perception) & $94.63 \pm 0.56$ & $98.55 \pm 0.20$ \\
Xception \cite{chollet2017xception} & $94.72 \pm 0.91$ & $98.35 \pm 0.28$ \\
\hline
\end{tabular}
\label{table:results_0}
\end{center}
\caption{Verification accuracies (\% with standard deviations) of different algorithms on LFW Image-Unrestricted settings. The upper half lists previous state-of-arts and the bottom half are the experimental results. The standard deviation for ArcFace algorithm was not found.}
\end{table}


\section{Conclusion}
In the present study, I proposed FusiformNet, a novel framework to extract discriminative facial features using a bottleneck autoencoder and a perceptual DNN. Validated on LFW face verification task, FusiformNet achieved significant performances on facial recognition with simple settings. Although there have been several approaches represented by special loss functions, image normalization and augmentation, and multi-input architectures, as far as I know, FusiformNet is the first framework to focus on and leverage the nature of discriminative facial features for their effective extraction. The experimental results confirmed that separately extracting general and local person-identifying features benefits recognition performance, although general or local facial features alone do not. Although FusiformNet is tested only on a verification task, it plausibly also applies to identification tasks.

There still are rooms for improvement, as the present study only tested the basic performance of FusiformNet under a simple condition. For example, the bottleneck width of the autoencoder, the architecture and the weight trainability of the autoencoder and the perceptual DNN, etc. can be tuned through further examination. In addition, like TP-GAN \cite{huang2017beyond}, features could be parallelly extracted by separate facial regions (eyes, mouth, nose, etc.). For more practical applications, FusiformNet can be easily combined with previous successful approaches like ArcFace loss \cite{deng2019arcface}, GAN-based face rotation, etc.

It is also notable that FusiformNet could be potentially applied to other DNN-based inferences. One potential example is psychological analyses of associations between certain personal attributes and facial features \cite{wang2018deep, wu2016automated, mehta2019recent}. If such associations are found both generally and locally on individuals’ faces, FusiformNet might be a better choice than mere ImageNet DNNs on estimating presences of such personal attributes. Needless to say, researchers---whether psychologists or engineers--- must pay great attention to ethical aspects and such a use of DNNs should be balanced between utility and ethical concerns, like other technologies and studies. Also, similar approaches might be also effective for recognition tasks involving other biometrics such as fingerprint and speaker recognition.

{\small
\bibliographystyle{ieeetr}
\bibliography{egbib}
}

\end{document}